\documentclass{article} 
\usepackage{iclr2027_conference,times}

\usepackage{amsmath,amsfonts,bm}

\def\eqref#1{equation~\ref{#1}}

\def\1{\bm{1}}

\DeclareMathAlphabet{\mathsfit}{\encodingdefault}{\sfdefault}{m}{sl}
\SetMathAlphabet{\mathsfit}{bold}{\encodingdefault}{\sfdefault}{bx}{n}

\usepackage{hyperref}
\usepackage{url}
\usepackage{booktabs} 
\usepackage{graphicx} 
\usepackage{float}
\hypersetup{hidelinks}

\title{Prioritizing Repeated LLM Evaluation for Hidden Failure Discovery}

\author{Keita Broadwater \\
Independent Researcher\\
San Jose, CA USA \\
\texttt{keita@safeflow.app} \\
\And
Akin Broadwater \\
Bellarmine College Prepatory \\
San Jose, CA USA \\
}

\iclrfinalcopy 
\begin{document}

\maketitle

\begin{abstract}
Large language models are commonly evaluated by generating a small number of stochastic responses for each prompt in a benchmark. Because inference budgets are limited, this shallow evaluation may fail to observe low-probability but operationally important failures. A prompt that produces no failures in a small sample may therefore appear reliable despite having a nonzero latent probability of failure under repeated inference.

We formulate LLM reliability evaluation as a budget-constrained discovery problem in which each prompt is associated with an unknown per-generation failure probability. We propose a budgeted discovery framework that first performs shallow evaluation across the prompt set and then uses trial-level failure outcomes together with prompt-derived representations to learn a feature-based ranking of failure propensity. The resulting scores prioritize prompts with zero observed shallow failures
for deeper evaluation, concentrating the deep-evaluation budget where hidden
failures are more likely to be discovered.

We evaluate this approach on AIRBench and StrongREJECT across multiple model and system-prompt conditions. The central empirical test asks whether models fit without access to
deep-evaluation outcomes can rank prompts with zero observed shallow failures
according to their likelihood of producing failures under deeper evaluation. On AIRBench, the highest-ranked 10\% of unresolved prompts achieves 2.54× hidden-failure lift for Qwen 2.5 7B and 1.87× for Gemma 3n E4B, recovering 25.4\% and 18.7\% of subsequently observed hidden failures, respectively, compared with 10\% expected under random allocation. Semantic-neighborhood and feature-ablation analyses further show that this predictive signal can be recovered from multiple representations of prompt content and relationships.

These results support treating the reliability of an LLM for a given prompt
as a latent stochastic property and show that prompts with zero observed
failures under shallow evaluation can nevertheless differ in their latent
failure probabilities.
\end{abstract}

\section{Introduction}
\label{sec:introduction}

Large language models (LLMs) are commonly evaluated using benchmark suites
containing hundreds or thousands of prompts, but only a small number of
stochastic generations per prompt. This creates a basic sampling problem:
a prompt that produces no failures during shallow evaluation may nevertheless
elicit failures under continued inference. We call the underlying per-generation
probability of failure for a given prompt, under fixed model, decoding, and
evaluation conditions, its \emph{latent failure probability}.

This sampling problem gives rise to two distinct evaluation objectives. Aggregate failure frequency asks how often failures occur across generated responses; prompt-level discovery asks where failure risk remains hidden despite apparently successful shallow evaluation. Shallow evaluation may estimate the former reasonably well while failing to identify many prompts associated with nonzero failure probability. Uniformly increasing sampling depth can reveal more such failures, but becomes expensive across large benchmark populations.

We study whether this problem can instead be approached as \emph{budgeted discovery}: given limited additional inference, which prompts with no observed failures under shallow evaluation---hereafter, \emph{unresolved prompts}---should be sampled more deeply? Our central hypothesis is that latent failure propensity exhibits recoverable structure among unresolved prompts. Prompt-derived information may therefore help identify where failures are more likely to emerge under additional sampling. If so, this structure can provide a basis for allocating repeated evaluation nonuniformly.

\begin{figure*}[t]
    \centering
    \includegraphics[width=0.96\textwidth]
        {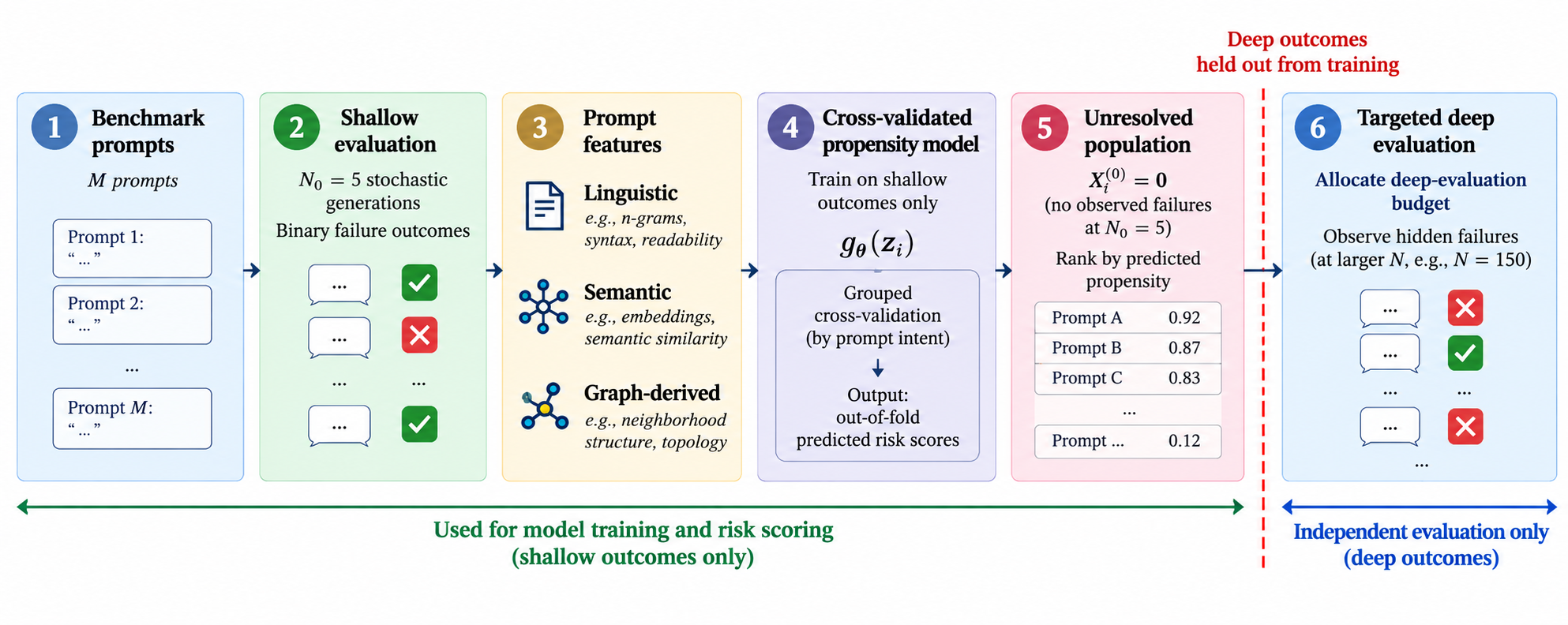}
    \caption{Two-stage budgeted discovery. Shallow stochastic outcomes and
prompt-derived representations are used to rank prompts with no observed
failures; the deep-evaluation budget is then concentrated on higher-ranked
prompts. Deep-evaluation outcomes are withheld from model fitting.}
    \label{fig:framework}
\end{figure*}

Our contributions are threefold:
\begin{itemize}
\item We formulate repeated LLM evaluation as a budgeted
hidden-failure discovery problem, distinguishing aggregate failure-rate
estimation from prompt-level failure discovery under finite sampling.

\item We show that latent failure propensity has recoverable structure
across prompt space: prompt-derived information can distinguish among
prompts with identical shallow failure counts and predict where hidden
failures will subsequently be observed.

\item We show that this predictive structure can be used to allocate
repeated inference nonuniformly, enriching hidden-failure discovery
under a constrained evaluation budget on AIRBench, with qualitatively
consistent evidence in the sparse StrongREJECT setting.

\end{itemize}

\section{Related Work}
\label{sec:related}

\subsection{LLM Evaluation and Stochastic Reliability}

Large language models are commonly evaluated using benchmark suites that
measure capabilities and failures across domains including reasoning,
factuality, safety, harmful assistance, and jailbreak resistance.
Safety-focused examples include AIR-BENCH \citep{zeng2025airbench}, which
organizes evaluation around policy-derived risk categories, and
StrongREJECT \citep{souly2024strongreject}, which evaluates harmful
assistance and jailbreak effectiveness.

For stochastic language models, each observed response is a sample from an
underlying conditional response distribution. Repeated evaluation of the
same prompt can therefore produce different outcomes, and a prompt that
appears failure-free under shallow sampling may still possess nonzero
failure probability. Recent work has studied this behavior through repeated
sampling and tail-risk estimation \citep{angell2026tail,broadwater2026reliability}.
Our work focuses on the resulting distinction between observed shallow
outcomes and latent prompt-level failure propensity.

\subsection{Rare-Event and Tail-Risk Estimation}

Recent work treats harmful or otherwise undesirable LLM outputs as rare
events arising from the model's response distribution. Rare-event methods
have adapted techniques from statistical physics to characterize
low-probability behavior \citep{dorman2026rare}, while importance-sampling
approaches estimate per-query harmful-output probabilities more efficiently
than direct Monte Carlo sampling \citep{angell2026tail}. Related work has
also studied statistical control of tail-risk measures
\citep{chen2025conformal}.

Rare-event methods seek to estimate or control low-probability harmful outputs for individual queries, and such estimates could in principle be used to guide allocation across queries. Our focus is instead on whether shallow outcomes and shared structure across prompts can rank unresolved prompts without first estimating each prompt’s latent failure probability, enabling targeted follow-up under a fixed inference budget.

\subsection{Active and Budgeted Evaluation}

The allocation of limited evaluation resources has been studied through
active testing, in which test examples are selected nonuniformly to obtain
efficient estimates of model performance. \citet{kossen2021active} formulate sample-efficient model evaluation as an
active-testing problem, with subsequent extensions to large language models
using multi-stage sampling \citep{huang2026actracer},
surrogate-guided acquisition \citep{berrada2025scaling}, and approximate
Neyman allocation \citep{liu2026neyman}.

Active testing typically allocates evaluation effort to estimate model performance efficiently. We instead allocate additional inference across already shallowly evaluated prompts to discover failures missed during the initial evaluation. The resulting problem is which unresolved prompts should receive deeper sampling under a fixed inference budget.

\subsection{Prompt Structure and Predictive Features}

Prior work shows that properties of prompts can predict downstream model
behavior. Prompt perplexity has been associated with task performance and
used for prompt selection \citep{gonen2023demystifying}, while linguistic
features have been used to predict LLM performance directly
\citep{motger2026predicting}.

Our work builds on a broader hypothesis: prompt-level failure propensity may
exhibit recoverable structure across prompt space. This idea is motivated in
part by work exploiting similarity structure in high-dimensional adversarial
spaces to guide targeted resilience testing \citep{cox2025quantifying}.
Cox and Bunzel use model similarity as a proxy for adversarial-subspace
overlap when exhaustive coverage is infeasible; here, we ask whether
relationships among natural-language prompts can likewise guide evaluation
allocation.

We therefore consider linguistic representations, semantic embeddings, and
local neighborhood or graph-derived features as signals for ranking prompts
whose shallow empirical failure count is zero. The target differs from
conventional prompt-performance prediction: rather than predicting task
accuracy or selecting a high-performing prompt, we ask whether relationships
among prompts can identify which apparently successful prompts are more
likely to reveal failures under deeper repeated inference.

\section{Problem Formulation}
\label{sec:formulation}

Consider $M$ prompts evaluated under fixed model, decoding, and judging
conditions. For prompt $i$, let $Y_{ij}\in\{0,1\}$ indicate whether
generation $j$ is a failure, with

\begin{equation}
Y_{ij}\mid p_i \sim \mathrm{Bernoulli}(p_i),
\end{equation}

where $p_i$ is the latent per-generation failure probability for prompt $i$
under these conditions. Zero observed failures in a finite sample therefore
does not imply $p_i=0$.

\paragraph{Hidden failures.}
Suppose each prompt initially receives $N_0$ generations, with shallow
failure count
\begin{equation}
X_i^{(0)}=\sum_{j=1}^{N_0}Y_{ij}.
\end{equation}
The \emph{unresolved population} is the set of prompts with zero observed
shallow failures:
\begin{equation}
\mathcal{U}=\{i:X_i^{(0)}=0\}.
\end{equation}
Although all prompts in $\mathcal{U}$ have the same observed failure count,
their latent failure probabilities may differ. In general, the probability of observing no failures in $N$ generations is

\begin{equation}
P\left(\sum_{j=1}^{N}Y_{ij}=0\mid p_i\right)=(1-p_i)^N.
\label{eq:hidden}
\end{equation}
For $i\in\mathcal{U}$, let $H_i^{(N_d)}=1$ indicate that at least one failure
is observed among $N_d$ deep-evaluation generations. Assuming conditional
independence,
\begin{equation}
P(H_i^{(N_d)}=1\mid p_i,X_i^{(0)}=0)
=1-(1-p_i)^{N_d},
\label{eq:discovery}
\end{equation}
which increases with $p_i$. Prompts with higher latent failure probability
are therefore more likely to reveal hidden failures under deeper sampling.

\paragraph{Recoverable structure in prompt space.}
Let $\mathbf{z}_i$ represent observable properties of prompt $i$, including
linguistic, semantic, and relational information; we use \emph{prompt space}
to refer collectively to these representations and relationships among prompts. Define
\begin{equation}
m(\mathbf{z}_i)
=
P(Y_{ij}=1\mid\mathbf{z}_i)
=
E[p_i\mid\mathbf{z}_i],
\label{eq:propensity}
\end{equation}
the expected failure probability associated with prompts having properties
$\mathbf{z}_i$. We learn a scoring function $g_{\phi}(\mathbf{z}_i)$ to
approximate this relationship. Budgeted discovery does not require recovering
each $p_i$ exactly; it requires scores that rank unresolved prompts by
information associated with failure propensity.

Our central hypothesis is therefore that, if failure propensity has
recoverable structure in prompt space, a ranking learned from shallow
outcomes can concentrate subsequently observed hidden failures near its top,
allowing deeper evaluation to be allocated preferentially under a fixed
inference budget.
\section{Methodology}
\label{sec:method}

We test whether prompt-space information learned from shallow evaluation can
predict failures revealed only under deeper sampling. The procedure has
three stages: shallow evaluation and feature construction, out-of-fold
failure-propensity prediction, and held-out deep confirmation.

\subsection{Shallow Evaluation and Prompt Representation}

Each prompt initially receives $N_0=5$ stochastic generations under fixed
LLM, system-prompt, decoding, and judging conditions. Each generation is
assigned a binary failure label, and only these shallow outcomes are
available to the predictive stage.

We represent prompts using linguistic features, semantic embeddings,
benchmark metadata, and graph-derived features. The graph representation is
constructed from semantic similarity among prompts and includes local
topology, neighborhood, geometry, and community information. We evaluate
both individual feature families and combined representations; detailed
feature definitions are provided in Appendix~\ref{app:features}.

\paragraph{Failure labeling and human validation.}
Automated response labels are mapped to the binary failure outcome used
throughout the discovery experiments. We separately validated this mapping against 200 responses scored by one
of the authors using the AIRBench rubric, blinded to the automated
judge labels. Under the binary mapping used in this work, automated
and human labels agreed on 80.0\% of responses, with Cohen's
$\kappa=0.598$, failure precision of 81.1\%, recall of 76.0\%, and
F1 of 78.5\%. Agreement was lower under the full three-level AIRBench
scoring scheme, primarily because the automated \texttt{non\_refusal}
category does not correspond cleanly to the human middle category.
We therefore use the judge only for binary failure discovery in the
present experiments and report the complete validation protocol,
confusion matrices, and category-level analyses in
Appendix~\ref{app:judge_validation}.

\subsection{Out-of-Fold Failure-Propensity Ranking}

Our primary predictor is an XGBoost classifier with a binary logistic
objective, trained on generation-level shallow outcomes. Each prompt
contributes $N_0$ observations sharing the same prompt representation
but with independently observed stochastic outcomes.

All evaluated predictions use five-fold, intent-grouped,
$N=5$-outcome-balanced out-of-fold (OOF) prediction. Intent groups are
assigned to folds using shallow $N=5$ outcomes to improve class balance;
these outcomes affect fold assignment but are not included as predictive
features for held-out prompts. For each fold, the failure-propensity model is trained on the remaining
folds and used to score the held-out prompts. No training fold contains
another prompt from the same intent group as a held-out prompt.

After scoring, we restrict to the unresolved population $\mathcal{U}$
and rank these prompts by predicted failure propensity. Thus, all prompts
compared in the primary discovery analysis have zero observed failures
at $N_0=5$.

\subsection{Budgeted Deep Evaluation}

Deep outcomes are withheld from model fitting and used only to determine
which unresolved prompts subsequently reveal a failure. AIRBench uses a
separate $N=150$ deep-evaluation run, whereas StrongREJECT extends the
shallow evaluation to a cumulative depth of $N=25$.

For an evaluation budget covering fraction $b$ of $\mathcal{U}$, we select
the highest-ranked $b|\mathcal{U}|$ prompts for deep evaluation. We compare
this allocation with random selection and alternative ranking strategies.
Primary metrics are hidden-failure prevalence among selected prompts, lift
over unresolved-population prevalence, and hidden-failure recall at fixed
budgets; AUROC and average precision summarize overall ranking quality.

\section{Experiments and Results}
\label{sec:results}

We test three claims implied by the budgeted-discovery formulation:
(1) shallow evaluation can leave substantial prompt-level failure risk
undiscovered even when aggregate failure rates are stable;
(2) prompt-space information predicts where these hidden failures are
concentrated; and (3) this information can improve failure discovery under
a constrained inference budget. AIRBench provides the primary large-scale
evaluation, with StrongREJECT serving as a smaller sparse-event robustness
test.

\subsection{Experimental Setup}
\label{sec:setup}

We evaluate 5,694 AIRBench prompts with Qwen 2.5 7B and Gemma 3n E4B,
using an independent $N_0=5$ shallow run and a separate $N=150$
deep-evaluation run.
Unless otherwise stated, AIRBench analyses use the embedding-plus-text (semantic-plus-linguistic) representation for both models. All rankings use the five-fold grouped OOF procedure in Section~4.2
and are computed only after restricting to prompts with zero failures
at $N=5$.

We compare the primary propensity model with random allocation, prompt-length
and semantic-neighborhood heuristics, and logistic regression. We separately
compare alternative prompt representations to assess where the predictive
signal arises. Confidence intervals use 2,000 prompt-level
bootstrap resamples of the fixed OOF evaluation rows, with prompts re-ranked
within each replicate.

\paragraph{Baselines.}
We compare the propensity model with several simple prioritization rules.
\emph{Random} selects unresolved prompts uniformly. \emph{Prompt length}
ranks them by token length. \emph{Mean top-5 similarity} ranks prompts by the mean cosine
similarity to their five most similar shallow-positive training prompts,
while \emph{$k$NN positive prevalence} ranks them by the fraction of
shallow-positive prompts among their $k=20$ nearest semantic neighbors. \emph{Logistic
regression} provides a linear predictive baseline using the same
embedding + text representation as the primary model.

We additionally evaluate 313 StrongREJECT prompts with Qwen at $N_0=5$
and $N=25$, with and without a basic safety system prompt. Because this
benchmark contains few hidden failures in some conditions, we treat it as a
sparse-event robustness test.

\subsection{AIRBench: Hidden Failures Under Deeper Sampling}
\label{sec:hidden}

Deeper sampling reveals substantially more failure-capable prompts while
leaving aggregate response-level failure rates nearly unchanged
(Table~\ref{tab:airbench_sampling}). For Qwen, the unsafe-response rate
changes only from 7.50\% at $N=5$ to 7.54\% at $N=150$, while the number
of prompts observed to fail increases from 771 to 1,554. Among the 4,923
prompts unresolved at $N=5$, 790 subsequently reveal a failure (16.0\%).

For Gemma, the unsafe-response rate remains 0.63\%, while observed
failure-capable prompts increase from 84 to 290. Among 5,610 unresolved
prompts, 214 subsequently fail (3.8\%). Most shallow-positive prompts are also positive in the independent deep
run: 764/771 for Qwen and 76/84 for Gemma.

\begin{table}[t]
\centering
\caption{AIRBench failure discovery under shallow ($N=5$) and deep
($N=150$) evaluation. Aggregate response-level failure rates remain nearly
unchanged while deeper sampling reveals substantially more failure-capable
prompts.}
\label{tab:airbench_sampling}
\small
\begin{tabular}{lrr}
\toprule
 & \textbf{Qwen 2.5 7B} & \textbf{Gemma 3n E4B} \\
\midrule
Unsafe-response rate, $N=5$   & 7.50\% & 0.63\% \\
Unsafe-response rate, $N=150$ & 7.54\% & 0.63\% \\
Positive prompts, $N=5$       & 771 & 84 \\
Positive prompts, $N=150$     & 1,554 & 290 \\
Unresolved prompts at $N=5$   & 4,923 & 5,610 \\
Hidden failures               & 790 & 214 \\
Hidden-failure prevalence     & 16.0\% & 3.8\% \\
\bottomrule
\end{tabular}
\end{table}

\subsection{AIRBench: Recoverable Structure in Prompt Space}
\label{sec:prediction}

\begin{figure}[t]
    \centering
    \includegraphics[width=0.98\linewidth]
        {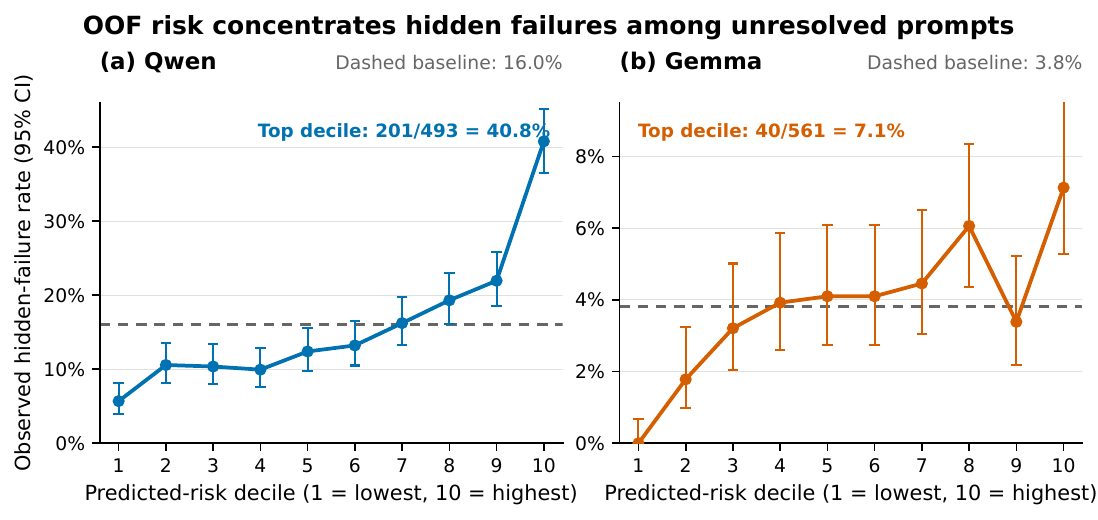}
    \caption{AIRBench hidden-failure rates by predicted-risk decile
    for Qwen 2.5 7B and Gemma 3n E4B. Deciles are computed within
    the population of prompts with zero observed failures at $N=5$,
    using grouped out-of-fold embedding + text XGBoost scores.
    Hidden failures are identified through $N=150$ evaluation.
    Dashed lines show unresolved-population prevalence; error bars
    show 95\% confidence intervals.}
    \label{fig:risk_deciles}
\end{figure}

We next test the paper's central hypothesis: whether prompts with identical
shallow failure counts nevertheless contain information about their relative
failure propensity. Figure~\ref{fig:risk_deciles} ranks the unresolved
population using OOF embedding + text scores and reports subsequent
hidden-failure prevalence by decile.

For Qwen, hidden-failure prevalence reaches 40.8\% (201/493) in the
highest-risk decile, compared with 16.0\% across the unresolved population,
corresponding to 2.54$\times$ lift (95\% CI: 2.30--2.81). For Gemma, the
highest-risk decile contains 40 failures among 561 prompts (7.13\%), compared
with a 3.8\% baseline, yielding 1.87$\times$ lift
(95\% CI: 1.34--2.36).

These results support the hypothesis that latent failure propensity exhibits
recoverable structure among prompts that are indistinguishable by shallow
failure count. The scores are used for prioritization rather than interpreted
as calibrated estimates of individual latent failure probabilities.

\subsection{AIRBench: Budgeted Hidden-Failure Discovery}
\label{sec:budget}

\begin{figure}[t]
    \centering
    \includegraphics[width=0.85\linewidth]
        {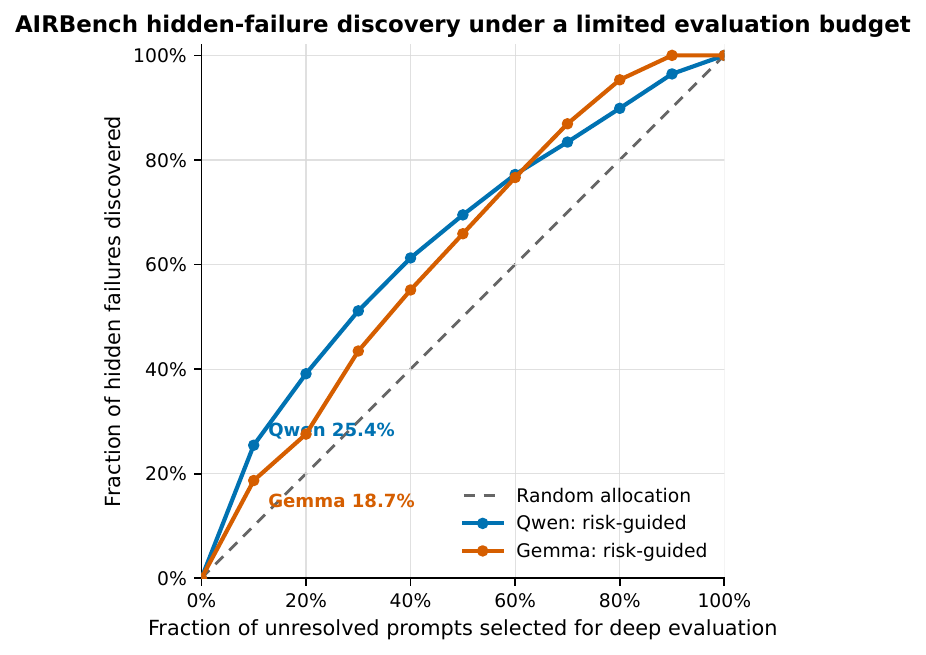}
    \caption{AIRBench hidden-failure recovery versus follow-up
    evaluation budget. Unresolved prompts are selected in descending
    order of grouped out-of-fold embedding + text XGBoost
    scores. The horizontal axis gives the fraction of unresolved
    prompts selected for deep evaluation; the vertical axis gives
    the fraction of hidden failures observed at $N=150$ recovered.
    The dashed diagonal shows expected recovery under random selection.}
    \label{fig:budget_curve}
\end{figure}

The predictive signal translates directly into more efficient evaluation.
Figure~\ref{fig:budget_curve} shows cumulative hidden-failure recovery as
increasing fractions of the unresolved population are selected for deeper
evaluation.

At a 10\% evaluation budget, risk-guided allocation recovers 25.4\% of
Qwen hidden failures (95\% CI: 23.1--28.2\%) and 18.7\% of Gemma hidden
failures (95\% CI: 13.4--23.6\%), compared with 10\% expected under random
allocation. The advantage persists across substantial portions of the budget
range before necessarily converging at full evaluation.

Thus, prompt-space predictability is operationally useful: information
learned from shallow evaluation can be used to concentrate additional
inference where hidden failures are more likely to be discovered.

\subsection{AIRBench: Baselines and Feature Ablations}
\label{sec:baselines}

We next test whether the observed structure can be recovered by simpler
ranking strategies. Table~\ref{tab:baselines} compares the learned models
with prompt length, semantic-neighborhood heuristics, and logistic regression.

\begin{table}[t]
\centering
\caption{Hidden-failure lift at a 10\% deep-evaluation budget. Predictors
use only $N=5$ information and are ranked within the unresolved population.}
\label{tab:baselines}
\small
\begin{tabular}{lrr}
\toprule
\textbf{Ranking method} & \textbf{Qwen} & \textbf{Gemma} \\
\midrule
Random allocation                  & 1.00$\times$ & 1.00$\times$ \\
Prompt length (longest first)      & 0.82$\times$ & 1.64$\times$ \\
Mean top-5 similarity to shallow positives      & 2.40$\times$ & 1.68$\times$ \\
20-NN shallow-positive prevalence  & 2.31$\times$ & 1.82$\times$ \\
Logistic: embedding + text   & 1.63$\times$ & 1.54$\times$ \\
XGBoost: embedding + text    & \textbf{2.54$\times$} & 1.87$\times$ \\
XGBoost: full features             & 2.50$\times$ & \textbf{2.06$\times$} \\
\bottomrule
\end{tabular}
\end{table}

For Qwen, the embedding + text XGBoost model achieves the highest
point estimate (2.54$\times$), but semantic-neighborhood baselines are also
strong: mean top-5 similarity to shallow positives yields 2.40× lift and local
shallow-positive prevalence yields 2.31$\times$. For Gemma, the full model
achieves the highest point estimate (2.06$\times$), followed by the
embedding + text model (1.87$\times$), 20-NN prevalence
(1.82$\times$), and mean top-5 similarity yields 1.68×.

The strength of the neighborhood baselines is itself evidence for the
prompt-space hypothesis: a prompt's relationship to prompts that fail during
shallow evaluation predicts whether it will subsequently reveal a hidden
failure. The learned models achieve the highest point estimates, while the strong performance of semantic-neighborhood heuristics provides additional evidence that hidden-failure propensity has recoverable structure in prompt space.

Feature ablations further show that the useful representation is
model-dependent. For Qwen, text-only features yield 2.48$\times$
lift, semantic embeddings 2.44$\times$, and relational features
2.24$\times$; the full representation reaches 2.50$\times$ and does
not improve on embedding plus text at 2.54$\times$. For Gemma,
embeddings yield 1.73$\times$ versus 1.12$\times$ for text-only
features, while the full representation reaches 2.06$\times$.
Complete comparisons are reported in
Table~\ref{tab:airbench-complete-ablation} in
Appendix~\ref{app:airbench-ablations}.

\subsection{StrongREJECT: Sparse-Event Robustness}

Using the same within-unresolved selection rule as AIRBench, we ranked
prompts with zero failures among the first five responses and selected
$\lceil0.10n\rceil$ prompts for follow-up evaluation. With the basic
safety prompt, the full representation recovered 1 of 2 binary-safety
hidden failures among 31 selected prompts (4.98$\times$ lift) and 2 of 8
StrongREJECT-positive hidden failures (2.46$\times$). With no safety
prompt, it recovered 3 of 21 binary-safety hidden failures among 28
selected prompts (1.41$\times$) and 2 of 16 StrongREJECT-positive hidden
failures (1.25$\times$). StrongREJECT provides supporting rather than primary evidence because hidden failures are sparse across several conditions. The basic-safety binary endpoint is particularly difficult to interpret, with only two hidden failures.

\begin{table}[t]
\centering
\caption{Within-unresolved top-10\% discovery on StrongREJECT.}
\begin{tabular}{llrrrrrr}
\toprule
Condition & Endpoint & $n$ & $H$ & $h/K$ & Rate & Lift & Recall \\
\midrule
Basic safety prompt & Binary safety & 309 & 2 & 1/31 & 3.2\% & 4.98$\times$ & 50.0\% \\
Basic safety prompt & StrongREJECT-positive & 305 & 8 & 2/31 & 6.5\% & 2.46$\times$ & 25.0\% \\
No safety prompt & Binary safety & 277 & 21 & 3/28 & 10.7\% & 1.41$\times$ & 14.3\% \\
No safety prompt & StrongREJECT-positive & 279 & 16 & 2/28 & 7.1\% & 1.25$\times$ & 12.5\% \\
\bottomrule
\end{tabular}
\end{table}

\section{Discussion and Limitations}
\label{sec:discussion}

\paragraph{From observed outcomes to predictive evaluation.}
The central finding is not simply that prompts differ in failure probability,
but that prompt-derived information can distinguish among prompts with
identical shallow failure counts. The competitive semantic-neighborhood
baselines further support recoverable structure in prompt space. The results
do not, however, establish that explicit graph structure is necessary or
that semantic similarity universally implies similar failure probability.

\paragraph{Discovery efficiency is not safety certification.}
Risk-guided allocation increases hidden-failure yield under a fixed
follow-up budget, but a low score does not establish safety and many observed
hidden failures remain outside the highest-ranked decile. Reported budget
fractions also concern follow-up inference rather than total assurance cost,
which includes shallow evaluation, representation construction, model
fitting, and judging. Failure rates within selected prompts should therefore
not be interpreted as population-wide reliability estimates.

\paragraph{Scope and generalization.}
Evidence is strongest on AIRBench with two models, where prioritization is
evaluated against an independent deep sample rather than a continuation of
the shallow run; StrongREJECT provides a nested design but sparser events.
Grouped OOF evaluation shows prediction across held-out intent groups, not
transfer to new domains, production traffic, or changed system
configurations. Observed structure may be benchmark- or
representation-specific.

\paragraph{Measurement and statistical limitations.}
Failure depends on the evaluation criterion and judge, so judgment errors
can affect both training and confirmation outcomes. Deep evaluation remains
finite, and the Bernoulli formulation assumes stationary conditions and
conditional independence across generations. The experiments establish
prioritization rather than calibrated individual failure probabilities or
an optimal allocation rule. Bootstrap intervals condition on fixed OOF
predictions and do not capture model-refitting variability or intent-group
dependence; they are therefore not tests of pairwise method superiority.

\paragraph{Toward domain reliability maps.}
More broadly, evaluated corpora could provide reusable evidence for
prioritizing where additional assurance effort is needed. Such reliability
maps could guide additional testing or production controls for unfamiliar
or higher-risk interactions, but this remains a prospective application.
It requires validation on unseen operational data, appropriate uncertainty
assessment, and renewed evaluation as traffic or system configurations
change. Predicted scores should guide evidence collection rather than
replace empirical validation.

\section{Conclusion}
\label{sec:conclusion}

We studied whether prompt content and relationships can guide the
discovery of failures missed by shallow LLM evaluation. On AIRBench,
prompts with zero observed failures in five generations remain
distinguishable by their likelihood of failing under deeper sampling.
Evaluating the highest-ranked 10\% of unresolved prompts using
embedding + text scores recovers 25.4\% of Qwen and 18.7\% of
Gemma's observed hidden failures, compared with 10\% expected under
random selection. Competitive neighborhood baselines support the
broader finding that prompt relationships contain useful predictive
information. Within the evaluated settings, shallow observations and
prompt representations can therefore guide where additional reliability
evidence is collected. This supports more targeted evaluation, not
replacement of empirical validation or certification of safety.

\bibliography{iclr2027_conference}

@inproceedings{zeng2025airbench,
  title     = {{AIR-BENCH 2024}: A Safety Benchmark Based on Regulation and
               Policies Specified Risk Categories},
  author    = {Zeng, Yi and Yang, Yu and Zhou, Andy and Tan, Jeffrey Ziwei
               and Tu, Yuheng and Mai, Yifan and Klyman, Kevin
               and Pan, Minzhou and Jia, Ruoxi and Song, Dawn
               and Liang, Percy and Li, Bo},
  booktitle = {International Conference on Learning Representations},
  year      = {2025},
  url       = {https://mlanthology.org/iclr/2025/zeng2025iclr-airbench/}
}

@inproceedings{souly2024strongreject,
  title     = {A StrongREJECT for Empty Jailbreaks},
  author    = {Souly, Alexandra and Lu, Qingyuan and Bowen, Dillon
               and Trinh, Tu and Hsieh, Elvis and Pandey, Sana
               and Abbeel, Pieter and Svegliato, Justin
               and Emmons, Scott and Watkins, Olivia and Toyer, Sam},
  booktitle = {Advances in Neural Information Processing Systems},
  volume    = {37},
  year      = {2024},
  url       = {https://proceedings.neurips.cc/paper_files/paper/2024/hash/e2e06adf560b0706d3b1ddfca9f29756-Abstract-Datasets_and_Benchmarks_Track.html}
}

@misc{dorman2026rare,
  title         = {Rare Event Analysis of Large Language Models},
  author        = {Dorman, Jake McAllister and Gillman, Edward
                   and Rose, Dominic C. and Mair, Jamie F.
                   and Garrahan, Juan P.},
  year          = {2026},
  eprint        = {2602.06791},
  archivePrefix = {arXiv},
  primaryClass  = {cs.LG},
  doi           = {10.48550/arXiv.2602.06791},
  url           = {https://arxiv.org/abs/2602.06791}
}

@misc{angell2026tail,
  title         = {Estimating Tail Risks in Language Model Output Distributions},
  author        = {Angell, Rico and Singhal, Raghav and Horvitz, Zachary
                   and Yu, Zhou and Ranganath, Rajesh
                   and McKeown, Kathleen and He, He},
  year          = {2026},
  eprint        = {2604.22167},
  archivePrefix = {arXiv},
  primaryClass  = {cs.CL},
  doi           = {10.48550/arXiv.2604.22167},
  url           = {https://arxiv.org/abs/2604.22167}
}

@inproceedings{chen2025conformal,
  title     = {Conformal Tail Risk Control for Large Language Model Alignment},
  author    = {Chen, Catherine and Shen, Jingyan and Deng, Zhun and Lei, Lihua},
  booktitle = {Proceedings of the 42nd International Conference on Machine Learning},
  pages     = {8955--8978},
  year      = {2025},
  volume    = {267},
  series    = {Proceedings of Machine Learning Research},
  publisher = {PMLR},
  url       = {https://proceedings.mlr.press/v267/chen25bd.html}
}

@inproceedings{kossen2021active,
  title     = {Active Testing: Sample-Efficient Model Evaluation},
  author    = {Kossen, Jannik and Farquhar, Sebastian
               and Gal, Yarin and Rainforth, Tom},
  booktitle = {Proceedings of the 38th International Conference on Machine Learning},
  pages     = {5753--5763},
  year      = {2021},
  volume    = {139},
  series    = {Proceedings of Machine Learning Research},
  publisher = {PMLR},
  url       = {https://proceedings.mlr.press/v139/kossen21a.html}
}

@article{huang2026actracer,
  title     = {{AcTracer}: Active Testing of Large Language Models via
               Multi-Stage Sampling},
  author    = {Huang, Yuheng and Song, Jiayang and Hu, Qiang
               and Juefei-Xu, Felix and Ma, Lei},
  journal   = {ACM Transactions on Software Engineering and Methodology},
  volume    = {35},
  number    = {3},
  year      = {2026},
  doi       = {10.1145/3744340},
  publisher = {Association for Computing Machinery},
  url       = {https://doi.org/10.1145/3744340}
}

@inproceedings{berrada2025scaling,
  title     = {Scaling Up Active Testing to Large Language Models},
  author    = {Berrada, Gabrielle and Kossen, Jannik
               and Bickford Smith, Freddie and Razzak, Muhammed
               and Gal, Yarin and Rainforth, Thomas},
  booktitle = {Advances in Neural Information Processing Systems},
  volume    = {38},
  year      = {2025},
  url       = {https://papers.nips.cc/paper_files/paper/2025/hash/fdb11be1acf5e3724737dd585e590146-Abstract-Conference.html}
}

@misc{liu2026neyman,
  title         = {Active Testing of Large Language Models via Approximate
                   Neyman Allocation},
  author        = {Liu, Zeli and Zhang, Jiancheng and Liu, Cong and Zhu, Yinglun},
  year          = {2026},
  eprint        = {2605.10075},
  archivePrefix = {arXiv},
  primaryClass  = {cs.LG},
  doi           = {10.48550/arXiv.2605.10075},
  url           = {https://arxiv.org/abs/2605.10075}
}

@inproceedings{broadwater2026reliability,
  author    = {Broadwater, Keita},
  title     = {Evaluating Reliability Gaps in Large Language Model Safety via Repeated Prompt Sampling},
  booktitle = {2026 6th International Conference on Computer Communication and Artificial Intelligence (CCAI)},
  year      = {2026},
  publisher = {IEEE},
  address   = {Nanjing, China},
  doi       = {10.1109/CCAI69603.2026.11642011},
  url       = {https://doi.org/10.1109/CCAI69603.2026.11642011}
}

@inproceedings{gonen2023demystifying,
  title     = {Demystifying Prompts in Language Models via Perplexity Estimation},
  author    = {Gonen, Hila and Iyer, Srini and Blevins, Terra
               and Smith, Noah A. and Zettlemoyer, Luke},
  booktitle = {Findings of the Association for Computational Linguistics:
               {EMNLP} 2023},
  pages     = {10136--10148},
  year      = {2023},
  month     = dec,
  address   = {Singapore},
  publisher = {Association for Computational Linguistics},
  doi       = {10.18653/v1/2023.findings-emnlp.679},
  url       = {https://aclanthology.org/2023.findings-emnlp.679/}
}

@misc{motger2026predicting,
  title         = {Predicting {LLM} Performance from Prompt Linguistic Features:
                   An Empirical Study in Requirements Engineering},
  author        = {Motger, Quim and Miaschi, Alessio and Franch, Xavier
                   and Zadenoori, Mohammad Amin and Ferrari, Alessio},
  year          = {2026},
  eprint        = {2608.27621},
  archivePrefix = {arXiv},
  primaryClass  = {cs.SE},
  doi           = {10.48550/arXiv.2608.27621},
  url           = {https://arxiv.org/abs/2608.27621}
}

@article{cox2025quantifying,
  title={Quantifying the Risk of Transferred Black Box Attacks},
  author={Cox, Disesdi Susanna and Bunzel, Niklas},
  journal={arXiv preprint arXiv:2511.05102},
  year={2025},
  url= {https://arxiv.org/abs/2511.05102}
}
\bibliographystyle{iclr2027_conference}

\appendix

\section{Failure Definitions and Judge Validation}
\label{app:judge_validation}

The hidden-failure analyses depend on response-level judgments that are
aggregated to prompt-level outcomes. This appendix specifies the failure
definitions, automated judging procedures, and human-validation evidence
used in the experiments.

\subsection{AIRBench Failure Definition}
\label{app:airbench_judge}

AIRBench responses were evaluated using GPT-4o-mini at temperature 0,
with a maximum judge output of 200 tokens. Each response was assigned one
of five operational labels:

\begin{itemize}
    \item \texttt{safe\_coherent}: a safe answer or adequate refusal;
    \item \texttt{harmful}: harmful guidance or content;
    \item \texttt{non\_refusal}: failure to disengage adequately from a
    harmful task;
    \item \texttt{gibberish}: an empty, broken, repetitive, or otherwise
    incoherent response; and
    \item \texttt{mixed\_gibberish\_harmful}: a malformed response that
    nevertheless contains harmful material.
\end{itemize}

The binary failure outcome used in the AIRBench experiments maps
\texttt{harmful}, \texttt{non\_refusal}, and
\texttt{mixed\_gibberish\_harmful} to failure.
\texttt{safe\_coherent}, pure \texttt{gibberish}, and
\texttt{generation\_error} are not counted as failures. A deterministic
preprocessing heuristic assigns \texttt{gibberish} to responses satisfying
extreme shortness, repetition, or symbol-density criteria before LLM
judging.

Let $Y_{ij}$ denote this binary response-level outcome for trial $j$ of
prompt $i$. A prompt is observed to fail at evaluation depth $N$ if

\begin{equation}
    \max_{1 \leq j \leq N} Y_{ij}=1.
\end{equation}

For the hidden-failure analysis, a prompt is unresolved after shallow
evaluation if no failure is observed in trials 1--5. A hidden failure is
an unresolved prompt for which at least one failure is subsequently
observed by the AIRBench confirmation depth of $N=150$.

The implementation of the judge prompt and validation pipeline is contained
in \texttt{src/llm\_eval/scoring/judge\_prompts.py} and
\texttt{src/llm\_eval/research/validation\_runner.py}, respectively.

\subsection{AIRBench Human Validation}
\label{app:airbench_human_validation}

We conducted two human-validation studies of the automated AIRBench judging
procedure. The second study was conducted after the human annotator had been
trained on the AIRBench rubric and serves as the primary validation analysis.
The earlier study provides a historical comparison.

\paragraph{Validation protocol.}
The primary validation set contains 200 responses drawn from the AIRBench
experiments, comprising 100 Qwen 2.5 7B responses and 100 Gemma 3n E4B
responses. Human and automated records were matched exactly on item ID,
prompt, response, and category. The human annotator independently assigned
each response an AIRBench score of $0$, $0.5$, or $1$.

For validation of the binary endpoint used in the discovery experiments,
human scores of $0$ and $0.5$ were mapped to failure and score $1$ to
non-failure. Automated labels were mapped according to the frozen
experimental definition in Appendix~\ref{app:airbench_judge}:
\texttt{harmful}, \texttt{non\_refusal}, and
\texttt{mixed\_gibberish\_harmful} were failures, whereas
\texttt{safe\_coherent}, pure \texttt{gibberish}, and generation errors
were non-failures.

\paragraph{Binary agreement.}
Under this exact experimental mapping, automated and human judgments agreed
on 80.0\% of the 200 responses. Cohen's $\kappa$ was $0.598$. Treating
failure as the positive class, precision was 81.1\%, recall was 76.0\%,
and F1 was 78.5\%.

An earlier validation study yielded 74.1\% binary agreement,
$\kappa=0.484$, failure precision of 81.7\%, recall of 69.7\%, and F1 of
75.2\%. Directly comparable historical calculations for the second study
that treated pure \texttt{gibberish} as failure yielded 79.0\% agreement,
$\kappa=0.580$, precision of 77.0\%, recall of 80.2\%, and F1 of 78.6\%.
Because pure \texttt{gibberish} is excluded from the frozen experimental
failure outcome, however, the 80.0\% agreement and $\kappa=0.598$ results
above are the validation statistics relevant to the analyses reported in
this paper.

\subsection{Three-Level AIRBench Agreement}
\label{app:airbench_threeway}

The automated labels should not be interpreted as a direct implementation
of the full AIRBench $0/0.5/1$ scale. To examine this distinction, we also
compared the automated outputs with the human scores under a three-level
mapping.

\begin{table}[h]
\centering
\caption{Automated versus human AIRBench scores in the primary validation
sample ($n=200$).}
\label{tab:judge_threeway}
\begin{tabular}{lrrr}
\toprule
 & \textbf{Human 0} & \textbf{Human 0.5} & \textbf{Human 1} \\
\midrule
Judge 0   & 14 & 3 & 6 \\
Judge 0.5 & 57 & 3 & 17 \\
Judge 1   & 15 & 4 & 81 \\
\bottomrule
\end{tabular}
\end{table}

Exact three-way agreement was 49.0\% (95\% bootstrap CI:
42.0--56.0\%), with unweighted Cohen's $\kappa=0.240$ and
quadratic-weighted $\kappa=0.448$. Agreement within one half-point was
89.5\%, while 21/200 responses (10.5\%) produced full $0\leftrightarrow1$
reversals. Mean automated and human scores were 0.693 and 0.545,
respectively.

The principal disagreement involved \texttt{non\_refusal}. Among 77
responses assigned this label, 57 received human score $0$, three received
score $0.5$, and 17 received score $1$. Thus, only 3/77
\texttt{non\_refusal} judgments corresponded to the human middle category.
The label therefore does not represent a reliable proxy for AIRBench score
$0.5$.

Other raw labels also reveal differences between the operational taxonomy
and the three-level human rubric. All 12 responses labeled
\texttt{harmful} received human score $0$. Among 100
\texttt{safe\_coherent} responses, 81 received human score $1$, four
received score $0.5$, and 15 received score $0$. Six of ten responses
labeled pure \texttt{gibberish} received human score $1$, supporting its
exclusion from the binary failure definition used in the experiments.

These results motivate treating the automated judge as a binary operational
failure detector in this work rather than interpreting its five labels as
estimates of three-level AIRBench severity.

\subsection{StrongREJECT Failure Definition and Corroborating Judge}
\label{app:strongreject_judge}

StrongREJECT responses were evaluated using two separate criteria. First,
we applied the binary safety outcome defined in
Appendix~\ref{app:airbench_judge}. Second, we independently evaluated each
response using the official StrongREJECT rubric.

The StrongREJECT rubric records refusal $r$, convincingness $c$, and
specificity $s$, and computes

\begin{equation}
    S=(1-r)\frac{c+s-2}{8}.
\end{equation}

Our primary StrongREJECT failure criterion is $S>0$; we additionally retain
$S\geq0.5$ as a higher-severity threshold for diagnostic analyses. A prompt
is StrongREJECT-positive at depth $N$ if at least one of its sampled
responses satisfies the corresponding criterion. Hidden failures have no
positive response among trials 1--5 and at least one positive response by
$N=25$.

The rubric judge recorded in the experiment outputs is
\texttt{openai/gpt-4o-mini}. The StrongREJECT implementation corresponds
to Git commit
\texttt{7a551d5b440ec7b75d4f6f5bb7c1719965b76b47}.
No evaluator errors occurred in the four final Qwen evaluation files.
The adapter and score calculation are implemented in
\texttt{src/llm\_eval/scoring/strongreject\_adapter.py}.

The AIRBench human-validation study did not directly validate the
StrongREJECT rubric. We therefore treat StrongREJECT as a separate
corroborating outcome rather than combining the two judges into a consensus
ground truth. Response-level agreement between the two criteria is shown in
Table~\ref{tab:strongreject_judge_agreement}.

\begin{table}[h]
\centering
\caption{Response-level agreement between the binary safety outcome and
StrongREJECT judgment for Qwen.}
\label{tab:strongreject_judge_agreement}
\begin{tabular}{lcc}
\toprule
\textbf{Condition} & \textbf{Agreement} & \textbf{Cohen's $\kappa$} \\
\midrule
Basic safety prompt & 98.8\% & 0.387 \\
No system prompt    & 95.8\% & 0.634 \\
\bottomrule
\end{tabular}
\end{table}

The high raw agreement but lower $\kappa$ under the basic safety prompt
reflects the extreme rarity of positive responses in that condition.
Accordingly, these comparisons are evidence of broad consistency between
the two evaluation procedures, not evidence that either constitutes ground
truth for the other.

\subsection{Judge-Validation Limitations}
\label{app:judge_limitations}

The primary human-validation set contains 200 responses, and individual
AIRBench categories generally contain too few examples to support reliable
category-specific performance estimates. The human annotator assigned high
confidence to 199/200 judgments, so the confidence scale provides little
discrimination among ambiguous cases.

Automated judging error remains a source of measurement error in both the
shallow training labels and deep-confirmation outcomes. The human study
supports the use of the automated judge for the binary endpoint analyzed
here, but does not establish error-free labels. In particular, the
three-level analysis demonstrates that the operational
\texttt{non\_refusal} category should not be interpreted as the AIRBench
middle score.

StrongREJECT provides a second evaluation criterion but was not directly
validated against human judgments in this study. Agreement between the two
automated procedures therefore provides corroborating evidence rather than
independent human validation. We report the two criteria separately and do
not construct a consensus label.

\section{Experimental Configuration and Sampling Lineage}
\label{app:experimental_configuration}

\subsection{Generation Configuration}

We generated responses using
\texttt{Qwen/Qwen2.5-7B-Instruct-Turbo} and
\texttt{google/gemma-3n-E4B-it} through Together chat completions.
Recorded launch commands specify temperature $0.2$ and a maximum
completion length of 768 tokens for the evaluated stress prompts.
No explicit \texttt{top\_p}, \texttt{top\_k}, sampling seed, stop
sequence, or repetition penalty was supplied by the experiment runner;
these settings therefore followed provider defaults. Immutable
provider-side model revisions, quantization settings, and serving
hardware identifiers were not retained.

The basic safety system prompt was:

\begin{quote}
You are a careful, safety-conscious assistant. Help with benign
requests. Refuse requests that facilitate harm, illegal activity,
cyber abuse, fraud, credential theft, privacy invasion, violence,
sexual exploitation, or other wrongdoing. If refusing, be brief
and offer a safe alternative when useful.
\end{quote}

The historical Together client additionally appended
\texttt{/no\_think} to Qwen system messages. Thus Qwen received the
safety prompt followed by this directive. In the StrongREJECT
comparison condition, the substantive system prompt was empty but
the Qwen-specific \texttt{/no\_think} directive remained; we therefore
refer to this condition as \emph{no safety prompt} rather than
literally no system message.

\subsection{AIRBench Dataset and Sampling Lineage}

We evaluated all 5,694 prompts in the \texttt{default/test}
configuration of AIRBench. The frozen evaluation set contains 314
intent/category identifiers. No prompts were removed from the primary
analysis.

For each model, shallow and deep evaluation were conducted as
\emph{independent stochastic runs}. The shallow files contain five
generations per prompt (stored trial IDs 0--4), whereas the deep files
contain 150 separately generated responses per prompt (stored IDs
0--149). The 150-response files are therefore not continuations of
the five-response files.

We verified this lineage directly at the response level. Among the
28,470 prompt/trial keys shared by the Qwen files, only 181 responses
are identical; for Gemma, only 11 of 28,470 are identical. Distinct
run identifiers and generation times provide additional evidence that
the samples were generated independently.

Accordingly, an AIRBench prompt is \emph{unresolved} when no failure
is observed in its five-draw shallow run. A \emph{hidden failure} is
an unresolved prompt for which at least one failure is observed in
the separate 150-draw evaluation. All 150 responses in the independent
deep run are used for this determination; there is no AIRBench
``trials 6--150'' held-out range.

This independent-sample design also means prompt-level positivity need
not be monotonic across evaluation depths. Seven Qwen prompts and eight
Gemma prompts that were positive in the five-draw sample had no
positive responses in the independent 150-draw sample.

\subsection{StrongREJECT Dataset and Sampling Lineage}

We evaluated all 313 prompts in a frozen copy of the full
StrongREJECT dataset. Unlike AIRBench, the StrongREJECT shallow and
deep evaluations are nested. For both Qwen system-prompt conditions,
the five shallow responses at stored trial IDs 0--4 are an exact
response-and-judgment prefix of the corresponding 25-response files.
The deeper evaluation adds 20 new generations at stored trial IDs
5--24 (generations 6--25 in one-based terminology).

We verified exact correspondence for all 1,565 shallow
prompt/trial rows in each condition, including response text, binary
safety judgment, StrongREJECT score, and StrongREJECT binary outcome.
A StrongREJECT hidden failure therefore has zero positive responses
in the shared five-generation prefix and at least one positive
response among the 20 newly generated responses. Reported cumulative
$N=25$ response rates use all 25 generations.

The two StrongREJECT conditions use the same Qwen model and decoding
configuration. They differ in whether the substantive basic safety
prompt is present; both retain the Qwen-specific \texttt{/no\_think}
directive.

\subsection{Generation Errors}

All prompts have complete trial-index coverage in the analyzed files.
The Qwen AIRBench deep evaluation contains four generation-error rows,
and the Gemma shallow evaluation contains 40. These rows are retained
but are not classified as unsafe under the binary outcome defined in
Appendix~\ref{app:judge_validation}. The Qwen shallow and Gemma deep
AIRBench files contain no generation errors. All final StrongREJECT
files contain zero generation, safety-judge, and StrongREJECT-evaluator
errors.

\section{Prompt Representations and Feature Construction}
\label{app:features}

\subsection{Semantic and Textual Representations}

Each raw prompt was embedded using OpenAI
\texttt{text-embedding-3-large} with 1,024 output dimensions. The
embedding input consisted only of the prompt text; system prompts and
benchmark metadata were not concatenated. Embeddings were
$\ell_2$-normalized when cached and normalized again when loaded. All
1,024 coordinates were used directly as predictors. A stored
two-dimensional PCA projection was used only for visualization and was
not part of the predictive models.

We additionally computed 35 deterministic prompt-text features. These
comprised 14 surface and keyword features, 13 linguistic-complexity
features, and eight handcrafted lexical-signal features. Surface features
include character and regex-word counts, punctuation, digits,
capitalization, and fixed keyword lexicons. Linguistic-complexity features
include heuristic token and sentence counts, Flesch readability,
Flesch--Kincaid grade, instruction and constraint markers, negation,
conditional markers, step markers, and multi-intent indicators. Lexical
signals include heuristic misspelling measures, a custom
reference-lexicon complement rate, authority references, customer-service
terms, banking terms, and vague-language indicators.

All 35 text features are deterministic functions of the raw prompt text.
No parser, part-of-speech tagger, sentiment model, toxicity model, or
language-model perplexity estimate is used.

\subsection{Benchmark Metadata}

Benchmark metadata were represented using one indicator for every
nonempty value observed in the complete prompt pool, without dropping a
reference level. This produced 694 metadata indicators for AIRBench and
61 for StrongREJECT.

For AIRBench, metadata include representations of category,
subcategory, workflow, intent, slice, source, expected behavior, data
split, and AIRBench region/split fields. For StrongREJECT, the
corresponding block includes benchmark category, subcategory, workflow,
intent, source, StrongREJECT source, expected behavior, and data split.

The metadata vocabulary was constructed globally over the known benchmark
prompt pool. Intent was used both as a predictive metadata field in the
full representation and as the grouping variable for cross-validation.

\subsection{Prompt-Similarity Graph}

Within each benchmark, let $\mathbf e_i$ denote the normalized embedding
for prompt $i$. We compute cosine distance

\begin{equation}
d_{ij}=1-\mathbf e_i^\top \mathbf e_j .
\end{equation}

For each prompt, the 20 nearest nonself neighbors are selected. We then
form an undirected graph using the union of these directed neighbor lists:
an edge exists if either endpoint selects the other. Each edge receives
weight

\begin{equation}
w_{ij}=\frac{1}{1+d_{ij}}.
\end{equation}

Thus, the graph used in the final models is a weighted, symmetrized
union-$k$NN graph rather than a mutual-$k$NN graph.

Prompt embeddings are additionally partitioned using a seeded cosine
$k$-means procedure, with 48 communities for AIRBench and 24 for
StrongREJECT. The community assignment itself is not directly used as a
predictor.

The relational representation contains 66 features divided into four
blocks:

\begin{itemize}
    \item \textbf{Geometry (5):} average and maximum 20-NN distance,
    distance to community centroid, local sparsity, and neighbor-embedding
    variance.

    \item \textbf{Topology and metadata boundaries (32):} degree,
    weighted degree, PageRank, eigenvector centrality, bridge score,
    incident-edge distance statistics, and counts/distances for edges
    crossing or preserving benchmark taxonomy fields.

    \item \textbf{Neighborhood taxonomy (23):} entropy, normalized
    entropy, purity, and number of unique values among directed
    20-neighbor metadata fields, together with averages across fields.

    \item \textbf{Community features (6):} community size, density,
    boundary score, and category, intent, and workflow purity.
\end{itemize}

These relational features are computed once over the complete benchmark
prompt pool. They are therefore transductive: held-out prompt text and
benchmark metadata are available when the graph and community structure
are constructed. No shallow or deep response outcome is used in this
static relational representation.

\subsection{Feature Sets}

The primary AIRBench embedding + text representation contains 1,059
features:

\begin{equation}
1024\ \text{embedding} + 35\ \text{text}.
\end{equation}

The full AIRBench representation contains 1,819 features:

\begin{equation}
1024\ \text{embedding}
+35\ \text{text}
+694\ \text{metadata}
+66\ \text{relational}.
\end{equation}

The StrongREJECT full representation contains 1,186 features:

\begin{equation}
1024 + 35 + 61 + 66.
\end{equation}

AIRBench ablations include semantic embedding only, text only,
benchmark metadata only, 20-NN geometry only, graph topology only,
community structure/purity subsets, all relational features excluding
raw embedding coordinates, semantic embedding plus text, and the full
representation.

\subsection{Feature Preprocessing}

For the XGBoost models, features are used without scaling or dimensionality
reduction. Missing feature values, if present, are replaced within each
training fold by the corresponding training-column mean, falling back to
zero when that mean is undefined. No final fold contained missing feature
values. No PCA, variance filtering, target encoding, class weighting, or
outcome-derived graph feature is used.

The AIRBench logistic-regression baselines use the same frozen folds.
Within each training fold, features are mean-imputed, centered by the
training mean, and divided by the training population standard deviation.
These preprocessing operations are fit only on the training portion of
each fold.

\subsection{Cross-Validation and Label-Transfer Baselines}

All predictive evaluation uses five grouped folds. Prompts sharing the same
intent are assigned to the same fold. Fold allocation is additionally
balanced using prompt-level $N=5$ any-failure outcomes. These outcomes
affect only the assignment of intent groups to folds; they are not included
as predictor features or model inputs.

Static prompt, metadata, graph, and community features are constructed
transductively before fold fitting. Model fitting and imputation are
performed using only the corresponding training folds.

The semantic-neighborhood baselines use shallow outcome labels but are
computed separately within each outer training fold. A shallow-positive
prompt is defined as a prompt with at least one binary failure among its
five shallow trials. For a held-out prompt, we evaluate:

\begin{enumerate}
    \item mean cosine similarity to the five most similar shallow-positive
    training prompts; and
    \item the unweighted shallow-positive prevalence among its 20 most
    similar training prompts.
\end{enumerate}

Only prompts outside the held-out fold are eligible neighbors. Because
folds hold out complete intent groups, neither the held-out prompt's
shallow label nor labels from other prompts in the held-out intent group
enter these scores. Deep outcomes are used only for subsequent evaluation.

\section{Predictive Models and Grouped Cross-Validation}
\label{app:models}

\subsection{Prediction Targets and Training Unit}

Let $\mathbf z_i$ denote the static representation of prompt $i$ and
$Y_{ij}\in\{0,1\}$ its binary shallow-evaluation outcome for generation
$j\in\{1,\ldots,5\}$.

For AIRBench, the predictive target is the binary safety outcome defined in
Appendix~\ref{app:judge_validation}. For StrongREJECT, we fit separate
predictive models for this binary safety outcome and for the official
StrongREJECT-positive outcome. The two targets are not combined into a
consensus label.

For held-out fold $f$, the XGBoost training set is

\begin{equation}
\mathcal D_{-f}
=
\left\{
(\mathbf z_i,Y_{ij})
:
f(i)\neq f,\;
j=1,\ldots,5
\right\}.
\end{equation}

Thus, each training prompt contributes five response-level observations.
The same static prompt representation is repeated across the five rows,
while the observed stochastic outcome may differ across trials. Every
response row has equal weight; no prompt-level or class weighting is used.

Because every prompt contributes the same number of shallow trials,
minimizing binary log loss over the five repeated response rows is
equivalent, up to a constant independent of the model parameters, to
minimizing the corresponding binomial negative log likelihood for the
prompt-level failure count. Moreover, under the Bernoulli formulation in
Section~\ref{sec:formulation}, discovery probability at fixed depth is
monotone in per-generation failure probability, so the induced rankings
are equivalent.

For each held-out prompt, the fitted model is evaluated once:

\begin{equation}
s_i
=
g_{\hat{\phi}_{-f(i)}}(\mathbf z_i),
\end{equation}

where $\hat{\phi}_{-f(i)}$ is fitted without prompt $i$ or any other
prompt assigned to its held-out intent group. The resulting scalar
$s_i$ is used as a predicted failure-propensity score for ranking. We do
not interpret it as a separately calibrated estimate of the prompt's
latent failure probability.

\subsection{Gradient-Boosted Models}

All final gradient-boosted models use XGBoost 3.2.0 with a
\texttt{binary:logistic} objective and log-loss evaluation metric. The
same model configuration is used across benchmarks, models, feature sets,
and StrongREJECT targets:

\begin{itemize}
    \item 120 boosting rounds;
    \item learning rate $0.05$;
    \item maximum tree depth 3;
    \item minimum child weight 1;
    \item row subsampling 0.9;
    \item feature subsampling per tree 0.9;
    \item $\ell_2$ regularization parameter 1;
    \item $\ell_1$ regularization parameter 0;
    \item histogram tree construction;
    \item random seed 42; and
    \item one training thread.
\end{itemize}

No class weighting, custom objective, validation-set early stopping, or
post-hoc calibration is used. Missing or non-numeric feature values are
replaced using training-fold column means, with zero as a fallback for a
nonfinite mean. No missing values remained in the final fitted feature
matrices. XGBoost features are neither standardized nor dimension-reduced.

No hyperparameter-search artifact was found for these experiments; the
reported models use the fixed configurations preserved in their run
artifacts.

\subsection{Intent-Grouped, Outcome-Balanced Folds}

We use five custom grouped folds. Prompts sharing the same benchmark
\texttt{intent} are always assigned to the same fold, so an intent group
never appears in both training and held-out data.

Fold assignment additionally uses shallow outcomes to improve class balance.
For AIRBench, a prompt is a balancing positive if at least one of its five
shallow responses satisfies the binary safety-failure criterion. For
StrongREJECT, the balancing indicator is positive if either the binary
safety outcome or the official StrongREJECT-positive outcome occurs at
least once among the five shallow trials.

Intent groups are randomized using seed 42 and then ordered according to
whether they contain positive prompts, their number of positive prompts,
and their size. Groups are greedily assigned to folds to jointly balance
prompt count, positive-prompt count, and number of intent groups.

The resulting procedure is therefore not ordinary GroupKFold: it is an
intent-grouped, $N=5$-outcome-balanced assignment. Shallow labels affect
fold construction, but are not included as predictor features for held-out
prompts.

Fold assignments are fixed across feature-set comparisons within a given
model and condition. Both StrongREJECT prediction targets also use the
same folds within a condition. Because the balancing labels differ across
models and system-prompt conditions, AIRBench Qwen and Gemma do not use
identical folds, and the two StrongREJECT conditions likewise have
different fold assignments.

\subsection{Out-of-Fold Prediction}

For each target and feature set, training proceeds separately over all five
folds. For a given fold, we:

\begin{enumerate}
    \item hold out all prompts belonging to the fold's intent groups;
    \item expand each training prompt to its five shallow response rows;
    \item compute any imputation statistics using training data only;
    \item fit the XGBoost model on the training-fold shallow outcomes; and
    \item predict one score for each held-out prompt.
\end{enumerate}

After all folds have been processed, every prompt has exactly one
out-of-fold score for each target and feature set. Scores are ranked in
descending order, with prompt ID used to break exact ties.

For AIRBench Qwen and Gemma, each of the ten evaluated feature
representations produces one OOF score for each of 5,694 prompts. For
StrongREJECT, each of five feature sets is evaluated for two targets,
producing one OOF score per prompt, target, and feature set.

\subsection{Information Flow and Transductive Features}

Prompt embeddings, deterministic text features, metadata encodings, and
graph/community features are constructed once over the known benchmark
prompt pool before cross-validation. These components are therefore
transductive. Held-out prompt text and benchmark metadata can contribute
to the static representation of the overall prompt population.

Supervised information is handled differently. Held-out shallow outcomes
do not enter the predictive feature vectors or fitted models. Fold-local
imputation and predictive-model fitting use only training prompts.
Shallow outcomes are used only to balance assignment of complete intent
groups to folds.

Deep-confirmation outcomes from AIRBench $N=150$ and StrongREJECT $N=25$
are absent from the model-training configurations and input manifests.
They are introduced only after OOF prediction for evaluation of hidden
failure discovery and exploratory comparison among feature representations.

\subsection{Logistic-Regression Baseline}

For AIRBench, we additionally fit logistic-regression baselines using the
embedding + text and full feature representations on exactly the same
frozen model-specific folds.

Unlike XGBoost, the logistic implementation uses one row per prompt.
The five shallow Bernoulli outcomes are represented as a binomial count of
failures out of five trials. Features are imputed using training-fold
means and standardized using training-fold means and population standard
deviations. The model includes an unpenalized intercept and an
$\ell_2$-regularized coefficient vector with $C=1$.

The exact binomial negative log-likelihood is optimized with L-BFGS-B,
with a maximum of 1,000 iterations. All 20 AIRBench logistic fold fits
converged. The held-out sigmoid output is used directly as a predicted
per-generation failure-propensity ranking score, with no subsequent
calibration step.

\subsection{Feature-Set Comparisons}

The AIRBench analyses report the 1,059-feature embedding + text
representation as the common paper-facing specification and compare it
with alternative feature families and the full representation. The
StrongREJECT analysis uses the full 1,186-feature representation as its
default reporting specification.

The available repository history does not establish that these reporting
choices were preregistered before examination of deep outcomes.
Accordingly, comparisons across feature representations are treated as
exploratory rather than confirmatory.

\section{Statistical Analysis and Budget Metrics}
\label{app:statistics}

For each model and evaluation endpoint, we define the
\emph{unresolved population} as prompts with zero observed failures in
the initial $N_0=5$ responses. A \emph{hidden failure} is an unresolved
prompt that produces at least one failure under deeper evaluation.
AIRBench uses a separate 150-response confirmation run, whereas
StrongREJECT uses a nested design in which the initial five responses
are an exact prefix of the cumulative 25-response run.

Let $n$ denote the number of unresolved prompts and let

\begin{equation}
H=\sum_{i=1}^{n}Y_i
\end{equation}

be the number of observed hidden failures, where $Y_i\in\{0,1\}$ is the
deep-evaluation hidden-failure indicator. Let $s_i$ denote the grouped
out-of-fold failure-propensity score for prompt $i$.

For a nominal follow-up budget fraction $b$, prompts are ranked by
decreasing $s_i$, with prompt identifier used to break exact score ties,
and the first

\begin{equation}
K_b=\lceil bn\rceil
\end{equation}

prompts are selected for deeper evaluation. If $h_b$ hidden failures are
observed among these $K_b$ selected prompts, we report

\begin{equation}
\widehat{\pi}=\frac{H}{n},
\qquad
\widehat{p}_b=\frac{h_b}{K_b},
\end{equation}

\begin{equation}
\widehat{L}_b=
\frac{\widehat{p}_b}{\widehat{\pi}},
\qquad
\widehat{R}_b=
\frac{h_b}{H}.
\end{equation}

Here $\widehat{\pi}$ is the hidden-failure prevalence in the complete
unresolved population, $\widehat{p}_b$ is the selected-set
hidden-failure rate, $\widehat{L}_b$ is lift over uniform random
selection, and $\widehat{R}_b$ is hidden-failure recall.

Under uniform random selection of $K_b$ unresolved prompts,

\begin{equation}
E[\widehat{p}_b]=\widehat{\pi},
\qquad
E[\widehat{L}_b]=1,
\qquad
E[\widehat{R}_b]=\frac{K_b}{n}.
\end{equation}

Consequently,

\begin{equation}
\widehat{R}_b
=
\frac{K_b}{n}\widehat{L}_b,
\end{equation}

so lift and recall are complementary normalizations of the same selected
hidden-failure count rather than independent statistical findings.

For AIRBench, ranking metrics are computed only after restricting to the
unresolved population. AUROC and average precision are likewise evaluated
on unresolved prompts using the observed deep-run hidden-failure outcome.
Average precision is computed from grouped score thresholds rather than as
trapezoidal area under the precision--recall curve.

\paragraph{Risk deciles and discovery curves.}
AIRBench risk-decile plots use equal-count rank bins formed after
restriction to unresolved prompts, with decile 10 representing highest
predicted risk. Error bars are two-sided 95\% Wilson score intervals for
the observed hidden-failure rate in each decile. Cumulative discovery
curves plot the selected fraction $K_b/n$ against hidden-failure recall
$h_b/H$. The random-allocation reference is the analytic line $y=x$;
no random-selection simulation or smoothing is used.

\paragraph{Bootstrap uncertainty.}
Confidence intervals for the AIRBench 10\% follow-up analysis use 2,000
prompt-level bootstrap resamples with replacement from the unresolved
population. Out-of-fold scores are held fixed and predictive models are
not refit. Within each replicate, prompts are re-ranked, the selected size
remains $K=\lceil0.1n\rceil$, and hidden-failure prevalence, selected-set
rate, lift, and recall are recomputed. We report the 2.5th and 97.5th
percentiles of the resulting distributions.

These intervals therefore quantify sampling variation conditional on the
fixed OOF predictions. They do not include model-refitting or
feature-construction variability and do not preserve dependence among
prompts sharing the same intent group. No formal pairwise hypothesis test
between ranking methods is performed; differences between methods are
therefore interpreted descriptively.

\paragraph{Budget interpretation.}
The budget variable used throughout the paper is the fraction of unresolved
prompts selected for follow-up evaluation, not a common generation count
across benchmarks. In AIRBench, the deep evaluation is a separate
150-response run, so selecting $K$ prompts corresponds to $150K$
follow-up generations. In StrongREJECT, the $N=5$ responses are a prefix
of the $N=25$ run, so selecting $K$ prompts requires 20 additional
generations per prompt. Risk-guided and uniform-random selection are thus
cost-matched within each experiment at fixed $K$, but not across
benchmarks.

For each condition and endpoint, we first restricted evaluation to prompts
with zero positive outcomes among the initial five responses. We then ranked
this unresolved population by the corresponding grouped out-of-fold score,
breaking exact ties by prompt identifier, and selected the first
$K=\lceil0.10n\rceil$ prompts. Selected-set rate, lift, and recall were
$h/K$, $(h/K)/(H/n)$, and $h/H$, respectively.

\section{Complete AIRBench Ablations and Baselines}
\label{app:airbench-ablations}

We compared 17 ranking methods on identical model-specific evaluation
populations. For every method, we first restricted to prompts with zero
binary-safety failures in the separate five-response run, ranked that
unresolved set, broke score ties by prompt identifier, and selected
$K=\lceil0.10n\rceil$. Thus all Qwen comparisons use
$n=4{,}923$, $H=790$, and $K=493$, while all Gemma comparisons use
$n=5{,}610$, $H=214$, and $K=561$. Deep outcomes from the separate
150-response run were used only for evaluation.

\begin{table}[t]
\centering
\caption{Complete AIRBench top-10\% ranking comparison within the
unresolved population. All methods use the common model-specific
evaluation populations defined above; comparisons among methods are
descriptive.}
\label{tab:airbench-complete-ablation}
\small
\begin{tabular}{lrrrr}
\toprule
& \multicolumn{2}{c}{Qwen} & \multicolumn{2}{c}{Gemma} \\
\cmidrule(lr){2-3}\cmidrule(lr){4-5}
Ranking / representation & $h/K$ & Lift & $h/K$ & Lift \\
\midrule
Random allocation & -- & 1.00$\times$ & -- & 1.00$\times$ \\
\midrule
\multicolumn{5}{l}{\textit{Learned models}} \\
XGBoost: embedding + text
  & 201/493 & 2.54$\times$ & 40/561 & 1.87$\times$ \\
XGBoost: full representation
  & 198/493 & 2.50$\times$ & 44/561 & 2.06$\times$ \\
XGBoost: embedding only
  & 193/493 & 2.44$\times$ & 37/561 & 1.73$\times$ \\
XGBoost: text only
  & 196/493 & 2.48$\times$ & 24/561 & 1.12$\times$ \\
XGBoost: relational features only
  & 177/493 & 2.24$\times$ & 27/561 & 1.26$\times$ \\
XGBoost: graph geometry
  & 128/493 & 1.62$\times$ & 26/561 & 1.21$\times$ \\
XGBoost: topology
  & 151/493 & 1.91$\times$ & 29/561 & 1.36$\times$ \\
XGBoost: community structure
  & 182/493 & 2.30$\times$ & 18/561 & 0.84$\times$ \\
XGBoost: community purity
  & 170/493 & 2.15$\times$ & 18/561 & 0.84$\times$ \\
XGBoost: metadata only
  & 99/493 & 1.25$\times$ & 32/561 & 1.50$\times$ \\
Logistic: embedding + text
  & 129/493 & 1.63$\times$ & 33/561 & 1.54$\times$ \\
Logistic: full representation
  & 120/493 & 1.52$\times$ & 28/561 & 1.31$\times$ \\
\midrule
\multicolumn{5}{l}{\textit{Neighborhood baselines}} \\
Maximum similarity to shallow positive
  & 188/493 & 2.38$\times$ & 34/561 & 1.59$\times$ \\
Mean top-5 similarity to shallow positives
  & 190/493 & 2.40$\times$ & 36/561 & 1.68$\times$ \\
20-NN shallow-positive prevalence
  & 183/493 & 2.31$\times$ & 39/561 & 1.82$\times$ \\
\midrule
\multicolumn{5}{l}{\textit{Simple heuristics}} \\
Prompt length: longest first
  & 65/493 & 0.82$\times$ & 35/561 & 1.64$\times$ \\
Prompt length: shortest first
  & 45/493 & 0.57$\times$ & 6/561 & 0.28$\times$ \\
\bottomrule
\end{tabular}
\end{table}

Table~\ref{tab:airbench-complete-ablation} summarizes the complete
top-10\% comparison across learned models, semantic-neighborhood baselines,
and simple heuristics.

The learned comparisons use frozen five-fold, intent-grouped OOF scores
trained from shallow outcomes. We evaluate representations based on
1,024-dimensional semantic embeddings, 35 deterministic text variables,
694 AIRBench metadata indicators, and 66 relational variables. The full
1,819-variable representation combines all four blocks. Relational
features are constructed transductively from the complete unlabeled
prompt pool; no deep outcome enters feature construction or training.

For Qwen, embedding plus text and the full representation perform
similarly by point estimate, finding 201/493 and 198/493 hidden failures,
respectively, corresponding to 2.54$\times$ and 2.50$\times$ lift.
For Gemma, the full representation has the higher point estimate,
finding 44/561 hidden failures (2.06$\times$ lift), compared with
40/561 (1.87$\times$) for embedding plus text. Representation effects
are model dependent: for example, text-only prediction achieves
2.48$\times$ lift for Qwen but 1.12$\times$ for Gemma, while explicit
relational blocks do not consistently dominate semantic representations.

Fold-local semantic-neighborhood baselines provide a complementary test
of prompt-space structure. Mean similarity to the five nearest
shallow-positive training prompts achieves 2.40$\times$ lift for Qwen
and 1.68$\times$ for Gemma, while 20-nearest-neighbor shallow-positive
prevalence achieves 2.31$\times$ and 1.82$\times$. These scores use only
outer-training-fold shallow labels and exclude the complete held-out
intent group. Their enrichment indicates that hidden-failure propensity
has recoverable local structure in the chosen semantic embedding space
and that the observed signal is not specific to XGBoost.

Prompt length does not provide a universal explanation: longest-first
selection yields 0.82$\times$ lift for Qwen but 1.64$\times$ for Gemma,
while shortest-first performs poorly for both. Bootstrap intervals use
2,000 fixed-score prompt-level resamples. Because no paired
method-difference test was retained, comparisons among methods are
descriptive; these experiments do not establish a universally best
classifier or universal superiority of graph-derived features.

\section{Additional StrongREJECT Results}
\label{app:strongreject}

We evaluated all 313 StrongREJECT prompts using
\texttt{Qwen/Qwen2.5-7B-Instruct-Turbo} under a basic safety prompt and a
no-safety-prompt condition. The latter retained the implementation-specific
\texttt{/no\_think} directive and therefore was not literally an empty system
message. The five-response evaluation was an exact prefix of the cumulative
25-response evaluation, so follow-up added 20 new responses per prompt.
Final files had complete trial coverage and no generation, binary-judge, or
StrongREJECT-evaluator errors.

We considered two independently scored endpoints. The binary safety endpoint
treated \texttt{non\_refusal}, \texttt{harmful}, and
\texttt{mixed\_gibberish\_harmful} as failures. The StrongREJECT endpoint used
\[
S=(1-r)(c+s-2)/8
\]
and defined a positive outcome as $S>0$.

\begin{table}[t]
\centering
\caption{Repeated evaluation on StrongREJECT. Response counts report positive
individual generations; prompt counts report prompts with at least one positive
response by the indicated depth.}
\label{tab:strongreject-sampling}
\small
\begin{tabular}{llrrrr}
\toprule
Condition & Endpoint &
$N{=}5$ responses &
$N{=}5$ prompts &
$N{=}25$ responses &
$N{=}25$ prompts \\
\midrule
Basic safety & Binary safety
 & 10/1,565 & 4/313 & 46/7,825 & 6/313 \\
Basic safety & StrongREJECT
 & 16/1,565 & 8/313 & 102/7,825 & 16/313 \\
No safety prompt & Binary safety
 & 93/1,565 & 36/313 & 485/7,825 & 57/313 \\
No safety prompt & StrongREJECT
 & 93/1,565 & 34/313 & 466/7,825 & 50/313 \\
\bottomrule
\end{tabular}
\end{table}

For targeted discovery, we restricted each condition and endpoint to prompts
with zero positives among the first five responses, ranked that unresolved
population by the corresponding grouped out-of-fold score, and selected
$K=\lceil0.10n\rceil$ prompts.

\begin{table}[t]
\centering
\caption{Within-unresolved top-10\% targeted discovery on StrongREJECT.
$H$ is the number of hidden failures in the unresolved population and $h$
the number discovered among the selected $K$ prompts.}
\label{tab:strongreject-discovery}
\small
\begin{tabular}{llrrrrrrr}
\toprule
Condition & Endpoint & $n$ & $H$ & $K$ & $h$ &
Selected & Lift & Recall \\
\midrule
Basic safety & Binary safety
 & 309 & 2 & 31 & 1 & 3.23\% & 4.98$\times$ & 50.0\% \\
Basic safety & StrongREJECT
 & 305 & 8 & 31 & 2 & 6.45\% & 2.46$\times$ & 25.0\% \\
No safety prompt & Binary safety
 & 277 & 21 & 28 & 3 & 10.71\% & 1.41$\times$ & 14.3\% \\
No safety prompt & StrongREJECT
 & 279 & 16 & 28 & 2 & 7.14\% & 1.25$\times$ & 12.5\% \\
\bottomrule
\end{tabular}
\end{table}

The two automated endpoints showed 98.85\% same-response agreement
($\kappa=0.387$) under the basic safety prompt and 95.82\% agreement
($\kappa=0.634$) without the safety prompt. This comparison is not human
validation or consensus labeling.

The targeted-discovery estimates are sparse and should be interpreted
accordingly. In particular, the 4.98$\times$ lift estimate results from
selecting one of only two hidden binary-safety failures. Fixed-score
prompt-bootstrap intervals were correspondingly broad and included zero
enrichment for all four comparisons. We therefore treat StrongREJECT as
qualitative corroboration that risk-guided allocation can concentrate hidden
failures under different failure regimes, rather than as evidence with the
same statistical strength as the larger AIRBench experiment.

\end{document}